# Generative Large Language Models Are All-purpose Text Analytics Engines: Text-to-text Learning Is All Your Need

Authors:  Cheng Peng, PhD[1]
Xi Yang, PhD[1,2]
Aokun Chen, PhD[1,2]
Zehao, Yu, MS[1]
Kaleb E Smith, PhD[3]
Anthony B Costa[3]
Mona G Flores[3]
Jiang Bian, PhD[1,2]
Yonghui Wu, PhD[1,2]

Affiliation of the authors:  [1]Department of Health Outcomes and Biomedical Informatics, College of Medicine, University of Florida, Gainesville, Florida, USA

[2]Cancer Informatics Shared Resource, University of Florida Health Cancer Center, Gainesville, Florida, USA

[3]NVIDIA, Santa Clara, California, USA

Corresponding author:  Yonghui Wu, PhD
Clinical and Translational Research Building
2004 Mowry Road, PO Box 100177
Gainesville, FL, USA, 32610
Phone: 352-294-8436
Email: yonghui.wu@ufl.edu

**ABSTRACT**

**Objective**

To solve major clinical natural language processing (NLP) tasks using a unified text-to-text learning architecture based on a generative large language model (LLM) via prompt tuning.

**Methods**

We formulated 7 key clinical NLP tasks as text-to-text learning and solved them using one unified generative clinical LLM, GatorTronGPT, developed using GPT-3 architecture and trained with up to 20 billion parameters. We adopted soft prompts (i.e., trainable vectors) with frozen LLM, where the LLM parameters were not updated (i.e., frozen) and only the vectors of soft prompts were updated, known as prompt tuning. We added additional soft prompts as a prefix to the input layer, which were optimized during the prompt tuning. We evaluated the proposed method using 7 clinical NLP tasks and compared them with previous task-specific solutions based on Transformer models.

**Results and Conclusion**

The proposed approach achieved state-of-the-art performance for 5 out of 7 major clinical NLP tasks using one unified generative LLM. Our approach outperformed previous task-specific transformer models by ~3% for concept extraction and 7% for relation extraction applied to social determinants of health, 3.4% for clinical concept normalization, 3.4~10% for clinical abbreviation disambiguation, and 5.5~9% for natural language inference. Our approach also outperformed a previously developed prompt-based machine reading comprehension (MRC) model, GatorTron-

MRC, for clinical concept and relation extraction. The proposed approach can deliver the "one model for all" promise from training to deployment using a unified generative LLM.

# INTRODUCTION

Solving multiple natural language processing (NLP) tasks with a unified model is an attractive promise of deep learning. Over the past decade, NLP researchers have successfully simplified feature engineering with vector representations [1] and pre-trained transformer models [2] that can be adapted to multiple downstream tasks. These advances have led to the development of large language models (LLMs), which could potentially fulfill the promise of "one model solving all NLP tasks." LLMs have dominated NLP in recent years, with generative LLMs particularly excelling at solving multiple NLP tasks through a unified sequent-to-sequence learning (i.e., text-to-text learning) [3] architecture, bypassing the need for generating vectors as an intermediate step. In the general domain, researchers have explored an encoder-decoder LLM, T5[4], to approach all NLP tasks as a "text-to-text" problem. However, the T5 model requires fine-tuning, meaning each subtask still needs a task-specific T5 variation, which is still one step away from the ideal of "one model solving all tasks".

This study seeks to apply the text-to-text learning architecture to formulate major clinical NLP tasks and solve them using a unified generative LLM through prompt tuning, thus fulfilling the "one model solving all tasks" promise. We formulate 7 major clinical NLP tasks as text-to-text learning problems, including (1) clinical concept extraction, (2) clinical relation extraction, (3) clinical abbreviation disambiguation, (4) natural language inference, (5) medication attribute filling, (6) clinical concept normalization, and (7) progress note understanding. We solved the 7 tasks using GatorTronGPT[5], a generative clinical LLM developed from scratch using a GPT-3 architecture and trained on 277 billion words of clinical and general English text. To solve the 7 tasks using a single unified model, we applied prompt-tuning with frozen LLMs, where the original parameters of GatorTronGPT remained unchanged during the tuning process. We used soft prompts[6] initialized by a bidirectional Long Short Term Memory (LSTM)[7] to instruct GatorTronGPT to generate accurate responses for different tasks. Our approach achieved state-of-the-art performance in 5 out of the 7 clinical NLP tasks, demonstrating the potential of generative clinical LLMs as versatile all-purpose text analytics tools capable of solving all major clinical NLP tasks. Soft prompting with frozen LLMs shows promise in facilitating the deployment of a unified generative LLM for diverse artificial intelligence (AI) applications.

## BACKGROUND

Understanding human language is a complex and challenging task. Researchers approach NLP using many intermediate subtasks, such as named entity recognition (NER) to identify important concepts, and word sense disambiguation (WSD) to determine the correct meaning of ambiguous words.[8] Similarly, clinical NLP has several major tasks including clinical concept extraction, clinical concept normalization, medical relation extraction, word sense disambiguation, natural language inference (NLI), medication attribute filling, and clinical note understanding. [8] Many clinical NLP open challenges have been organized to tackle these major challenges.[9–13]

Historically, NLP researchers developed different machine learning models for specific tasks. For example, sequence labeling models such as conditional random fields (CRFs) [14] are widely used for NER, and classification models such as support vector machines (SVMs)[15] for WSD. With the advent of deep learning, the field of NLP began to challenge the "one model per task" approach, aiming instead to solve multiple tasks with a unified model. Early deep learning methods for NLP involved training distributed word representations using shallow embedding algorithms such as word2vec[16], then integrating these embeddings into deep neural network architectures such as convolutional neural network [1,17,18] or recurrent neural network, particularly those implemented using LSTM [19–21], to solve various downstream tasks.

Not long after, the transformer [2,22] emerged, which is a neural network architecture implemented using a self-attention mechanism. A single pre-trained transformer model, like BERT [2]—an encoder-only transformer, can be adapted for multiple tasks through fine-tuning. While transformers have been widely used in many clinical NLP studies [23,24], adopting pre-trained transformers for different tasks typically requires adding task-specific layers and fine-tuning pre-trained parameters. This approach ultimately generates multiple task-specific models, deviating from the goal of "one model for all tasks".

Solving NLP tasks through text-to-text learning originates in the sequence-to-sequence learning architecture [3] proposed in the transformer model for machine translation. This model utilizes an "encoder" component to process the input sequence (i.e., source language) and a "decoder" component for generating the output sequence (i.e., target language). Recently, researchers have pre-trained very large transformer models with billions or even hundreds of billions of parameters using text data containing billions or even trillions of words.[25,26] These very large transformer models, known as LLMs, can be categorized into 3 main types: (1) encoder-only models, (2) decoder-only models, and (3) encoder-decoder models. LLMs using decoder-only and encoder-decoder transformers are "generative" as they can produce conversational responses, similar to humans.

Before 2021, encoder-only LLMs such as BERT dominated clinical NLP, widely used for tasks such as clinical concept extraction [23] and relation extraction [27]. Researchers have explored encoder-only transformers to unify NLP tasks as span extraction [28] or machine reading comprehension [29] tasks, which unified information extraction tasks such as clinical concept extraction and relation extraction. We previously developed GatorTron[30], a widely recognized encoder-only LLM in the clinical domain with up to 8.9 billion parameters. However, encoder-only LLMs require additional task-specific layers, which hinders their application in fulfilling the "one model for all tasks" promise.

More recently, generative LLMs such as the decoder-only GPT-3 model [25] and the encoder-decoder T5 model [4] have demonstrated strong text-to-text learning capabilities. In the general domain, Raffle *et al.* [4] first formulated multiple NLP tasks using text-to-text learning and solved them using T5-based LLMs. However, their approach still requires the fine-tuning of T5 parameters, resulting in multiple task-specific variations of T5 models, making it still a "one model per task" solution [4]. Agrawal *et al.* [31] recently examined few-shot learning ability of ChatGPT in solving multiple clinical information extraction tasks. In the clinical domain, we have developed GatorTronGPT [5], a generative clinical LLM trained from scratch using 277 billion words of clinical, biomedical, and general English text. We evaluated GatorTronGPT in solving biomedical relation extraction and biomedical question answering tasks. In a recent study [32], we explored soft prompts [6], i.e., a trainable vector that was added as a prefix to the input, and demonstrated that machines can learn soft prompts more effectively than human-composed hard prompts for clinical concept and relation extraction.

In this study, we seek to revisit the promise of "one unified model for all" and examine to what extent a single unified generative LLM can solve major clinical NLP tasks. Solving major clinical NLP tasks using a unified model is very attractive as it can significantly reduce training and deployment cost in real-world applications. This study is different from the previous study using the encoder-decoder T5 model [4] in three ways: (1) we explored a generative LLM based on a decoder-only architecture, specifically GPT-3[25], (2) we applied prompt tuning and fixed the parameters in the LLMs (i.e., frozen) during fine-tuning, whereas in [4], the T5 LLM was updated during fine-tuning, and (3) we adopted soft prompts instead of hard prompts composed by human experts as in [4]. The proposed approach in this study demonstrates the capability to solve all major clinical NLP tasks using a single unified generative LLM through soft prompting.

## METHODS

**Clinical NLP Tasks and Datasets**

We explored the following 7 major clinical NLP tasks:

***Clinical concept extraction*** aims to identify concepts with important clinical meanings (e.g., medications, treatments, adverse drug events). We used 2 benchmark datasets for evaluation: the 2018 n2c2 challenge (Track 2) dataset [33], focusing on the extraction of medication and adverse drug events (referred to as the drug-ADE dataset), and the 2022 n2c2 challenge (Track 2) dataset [34], focusing on social determinants of health (referred to as the SDoH dataset). The drug-ADE dataset consists of 505 discharge summaries from the Medical Information Mart for Intensive Care (MIMIC)-III database [35], annotated with 9 categories of clinical concepts (e.g., drug, drug attributes, ADEs). The SDoH dataset consists of 5 categories of SDoH concepts and 9 categories of SDoH-associated attribute concepts.

**Clinical concept normalization** is to standardize clinical concepts using standard concept identifiers, such as those defined in the Unified Medical Language System (UMLS). This task is typically approached through information retrieval and is solved using vector space models. [36] We used a disorder mention dataset developed for SemEval-2015 Task 14 for evaluation [37]. Specifically, the task is to detect disorder mentions and normalize them into a Concept Unique Identifier (CUI) within the UMLS/SNOMED-CT terminology. The dataset used is the ShARe corpus [38], which consists of 531 de-identified clinical notes (a mix of discharge summaries and radiology reports) selected from the MIMIC II clinical database [39].

*Clinical relation extraction* is to establish semantic relations among clinical concepts (e.g., drugs and adverse events). We used the drug-ADE dataset developed by 2018 n2c2 [33] and the SDoH dataset developed by 2022 n2c2 [34] for evaluation, where the drug-ADE dataset contains 8 categories of relations among drugs, drug-associated attributes, and ADEs, while the SDoH dataset consists of 28 categories of relations between SDoH concepts and SDoH-associated attributes.

*Clinical abbreviation disambiguation*, a special case of WSD, involves determining the correct meaning of ambiguous abbreviations in clinical narratives (e.g., AB for abortion). This task is typically solved using classification models. We utilized a widely used abbreviation dataset developed by the University of Minnesota (UMN), which consists of 37,500 instances, 75 unique abbreviations, and 351 senses.[40]

**Natural language inference (NLI)**, also known as recognizing textual entailment (RTE), involves determining if a given hypothesis can be inferred from a given premise, which is typically approached using classification models. We used MedNLI as the benchmark dataset, which was annotated by clinicians based on the medical history of patients [41]. There are three possible relationships between two sentences—a hypothesis and a corresponding premise, in NLI: *entailment (i.e., the hypothesis is logically follows from the premise)*, *contradiction (i.e., the hypothesis and premise are logically incompatible)*, and *neutral (i.e., the hypothesis is neither logically entailed nor contradicted by the premise)*.

**Medication attribute filling** is to determine the contextual attributes of medications documented in clinical narratives, which has been typically solved using multiple classification models and information extraction models. For evaluation, we used the Contextualized Medication Event Dataset (CMED), derived from the 2022 n2c2 challenge (Track 1), which focuses on the multi-dimensional context of medication changes [13]. This task is divided into two sub-tasks: event classification and context classification. The first sub-task aims to classify medication mentions into one of 3 categories: *Disposition*, *NoDisposition*, or *Undetermined*. The second sub-task aims to classify the contextual information for Disposition events across 5 orthogonal dimensions: *Action*, *Negation*, *Temporality*, *Certainty*, and *Actor*.

**Progress note understanding** aims to determine the causal relations between the "assessment section" and "plan section" of a progress note. For evaluation, we used the benchmark dataset developed by the 2022 n2c2 challenge (Track 3) [10], where a subset of 5,000 physician-written progress notes across 84 note types, representing the daily progress notes, are sampled from the MIMIC-III database. The annotated corpus contained 768 progress notes and 5,934 labels for predefined four relations: (1) Direct, (2) Indirect, (3) Neither, and (4) Not Relevant. The four relations corresponded to the providers' judgment on whether a diagnosis presented in the Plan Subsection was the primary reason for hospitalization (Direct), a secondary health problem or diagnosis to the main problem or diagnosis (Indirect), an issue that was not documented (Neither), or not a diagnosis or problem (Not Relevant).

**Generative LLMs**

We explored GatorTronGPT, a generative clinical LLM developed in our previous study [5]. GatorTronGPT was trained using a GPT-3 architecture and is available in two versions: one with 5 billion parameters and another with 20 billion parameters, using 277 billion words of text comprising (1) 82 billion words of clinical text from approximately 2 million patients at the University of Florida Health, and (2) 195 billion words of diverse general English text.[42]

**Prompting strategies**

We adopted soft prompting with frozen LLMs, which demonstrated good performance in our previous studies[29,32] based on an encoder-only LLM, GatorTron [30]. Soft prompts were used to instruct GatorTronGPT to generate correct responses for the 7 clinical NLP tasks. Specifically, the soft prompts were initiated as vectors of random values and optimized through backpropagation, while the LLM parameters remained frozen (i.e., not updated). This technique of using soft prompts, also known as prompt tuning, contrasts with the discrete 'hard' text-based prompts typically used by ChatGPT.[6] Given the input tokens, $\{x_1, x_2, ..., x_n\}$, where $n$ is the number of tokens, the GatorTronGPT model first embeds the tokens to forms a matrix $X_e \in R^{n \times e}$, with $e$ is the dimension of the embedding space. The soft prompts are represented as a parameter matrix $P_e \in R^{p \times e}$, where $p$ is the length of the prompt. The prompt parameter matrix $P_e$ is then concatenated with the embedded input $X_e$ to form a single matrix $[P_e; X_e] \in R^{(p+n) \times e}$, which then flows through the model layers. During training, the soft prompt parameters are updated while the LLM parameters are kept frozen. **Figure 1** compares the proposed soft prompting with frozen LLMs method with fully supervised training and pre-training/fine-tuning.

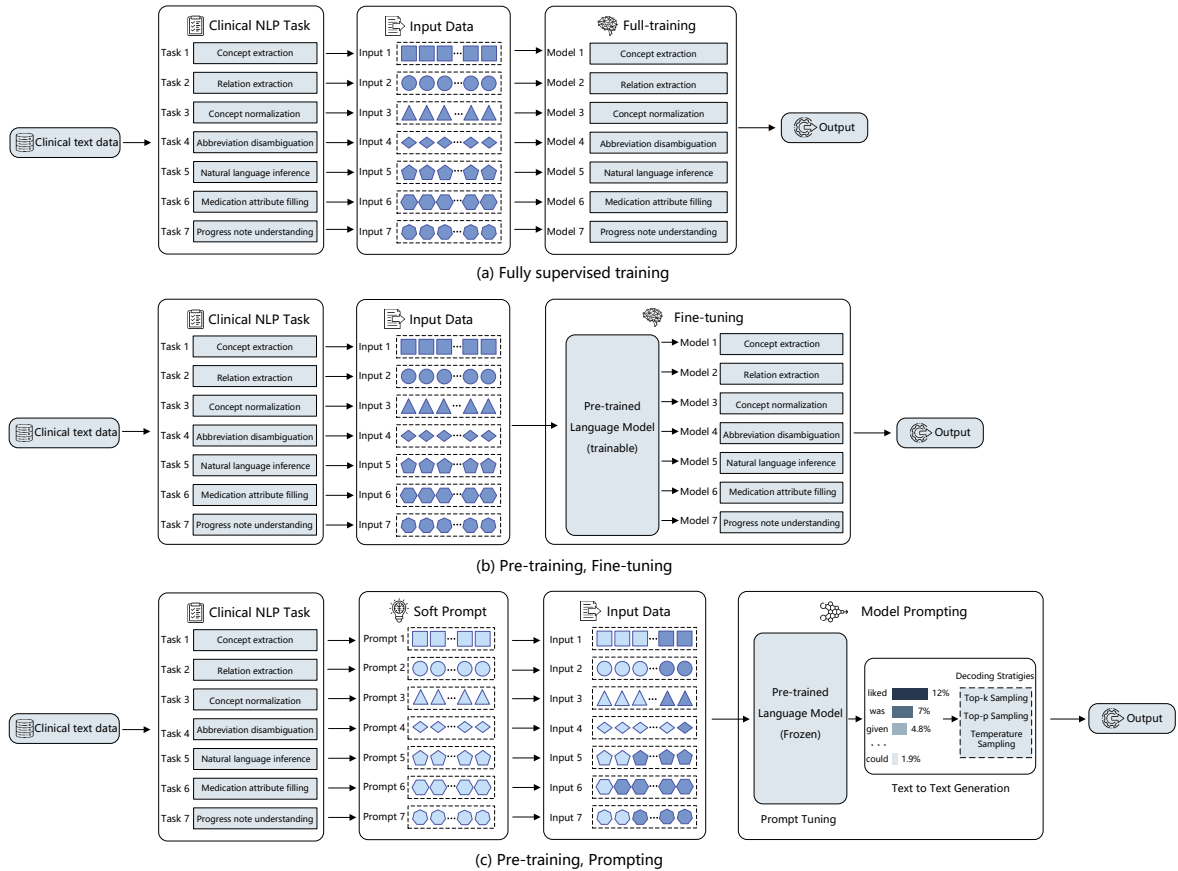

(a) Fully supervised training

(b) Pre-training, Fine-tuning

(c) Pre-training, Prompting

**Figure 1**. Three model tuning paradigms in clinical NLP tasks. Different shapes represent task-specific text data in the "Input Data" frame and different task-specific soft prompts in the "Soft Prompt" frame. **a**. Fully supervised learning. Seven individual models are trained from scratch on eight task-specific datasets. The outputs are obtained using task-specific objectives. **b**. Pre-training to fine-tuning. Seven individual models are fine-tuned on eight task-specific datasets based on a pre-trained model. The outputs are obtained using task-specific objectives. **c**. Pre-training to prompting. Seven individual soft prompts are trained on eight task-specific datasets based on a frozen pre-trained model. The outputs are obtained from a unified text generation module.

**Formulate clinical NLP tasks as text-to-text learning**

To solve all 7 clinical NLP tasks using a unified LLM, we reformatted all benchmark datasets into a text-to-text learning format, where both the input and output are in text form. **Table 1** provides examples for each of the 7 tasks. Specifically, we used "template" phrases to convert various forms of gold-standard annotations into natural language facts. For clinical concept normalization, we converted the UMLS concept CUIs into their official concept names to facilitate the text-to-text learning architecture. We concatenate a soft prompt as a prefix to the embeddings derived from the input text, denoted as [*prompt, input_text*], and then the annotated sample is constructed as [*prompt, input_text, target_text*]. During soft prompting, we use the gold standard *target_text* to optimize the soft prompts. During inference, we converted the generated *target_text* back to the format used by gold standard annotations for evaluation.

**Table 1**. Examples for formulating 7 clinical NLP tasks as text-to-text learning.

| Task | Input data | Annotation | Converted answer |
|---|---|---|---|
| Clinical concept extraction | 6. Colchicine 0.6 mg Tablet Sig: One (1) Tablet PO DAILY (Daily) as needed for Gout flare/pain. | Drug: Colchicin e<br>Strength: 0.6 mg<br>Form: Tablet, Tablet<br>Dosage: One (1)<br>Frequency: DAILY (Daily) as needed<br>Route: PO<br>Reason: Gout flare/pain | The extracted drug entity is **Colchicin e**; the extracted strength entity is **0.6 mg**; the extracted form entity is **Tablet**; the extracted dosage entity is **One (1)**; the extracted frequency is **DAILY (Daily) as needed**; the extracted route entity is **PO**; the extracted reason entity is **Gout flare/pain**. |
| Clinical relation extraction | Social History: Lives at [** Hospital6 3355 **], smoking 28 year pack hx, etoh remote, former IVDU (used once), [** Company 2318 **] bus driver for 18 yrs | (Lives, at [** Hospital6 3355 **], Living status-Type),<br>(smoking, 28 year pack hx, Tobacco-Amount),<br>(etoh, remote, Alcohol-Status),<br>(former IVDU, IVDU, Drug-Method),<br>(former IVDU, used once, Drug-Frequency), | The relation between "Lives" and "at [** Hospital6 3355 **]" is "**Living status-Type**"; the relation between "smoking" and "28 year pack hx" is "**Tobacco-Amount**"; the relation between "etoh" and "remote" is "**Alcohol-Status**"; the relation between "former IVDU" and "IVDU" is "**Drug-Method**"; the relation between "former IVDU" and "used once" is |

| | | (bus driver, for 18 yrs, Employment-Duration) | "**Drug-Frequency**"; the relation between "bus driver" and "for 18 yrs" is "**Employment-Duration**". |
|---|---|---|---|
| Clinical concept normalization | Past Medical History: Arthritis<br>carpal tunnel<br>shingles right arm 2000<br>needs right knee replacement<br>left knee replacement in [**2010**]<br>thyroidectomy 1978<br>cholecystectomy [**1981**]<br>hysterectomy 2001<br>h/o LGIB 2000-2001 after taking baby ASA 81 QOD<br>Social History:<br>Her husband died recently. | (arthritis, Arthritis, C0003864)<br>(carpal tunnel, Carpal Tunnel Syndrome, C0007286)<br>(shingles, Herpes zoster disease, C0019360)<br>(LGIB, Lower gastrointestinal hemorrhage, C0024050) | The normalized string of the disorder concept "**arthritis**" is "**Arthritis**"; the normalized string of the disorder concept "**carpal tunn**el" is "**Carpal Tunnel Syndrome**"; the normalized string of the disorder concept "**shingles**" is "**Herpes zoster disease**"; the normalized string of the disorder concept "**LGIB**" is "**Lower gastrointestinal hemorrhage**". |
| Clinical abbreviation disambiguation | \|CEA\|173\|175\|LABORATORY DATA\|PAIN: Negative. ADL STATUS: Energy: Low. Eating: She is eating well. Sleeping: She is sleeping well. Maintaining weight: Yes. LABORATORY DATA: Normal except for an elevated CEA at 6.1 but as the patient has been cutting back her smoking it has gone from 6.9 to 6.1. CHEMOTHERAPY/RADIATION THERAPY HISTORY: The patient has had no chemotherapy or hormone therapy. | carcinoembryonic antigen | The sense of the abbreviation "**CEA**" is "**carcinoembryonic antigen**". |
| Natural language inference | Premise: The patient was seen by his primary care physician after he had complained of a one-week history of dyspnea on exertion and jaw tightness.<br><br>Hypothesis: The patient has symptoms of a CHF exacerbation. | Entailment | The hypothesis that "The patient has symptoms of a CHF exacerbation" is **entailment** to the premise that "The patient was seen by his primary care physician after he had complained of a one-week history of dyspnea on exertion and jaw tightness". |
| Medication attribute filling | Context: "least moderate risk, positive stress test. After discussion with Dr. Camacho, our plan will be continue her on aspirin, beta-blocker, and a low-dose ACE inhibitor through her regimen. She refuses to take a cholesterol-lowering agent due to previous concerns with myalgias. She will be referred for cardiac catheterization within the next several days with a goal to better define her coronary anatomy for the possibility of percutaneous coronary intervention versus cardiac bypass surgery. | Event Classification:<br>(cholesterol-lowering agent, Disposition)<br><br>Context Classification:<br>(Action, Start)<br>(Negation, Negated)<br>(Temporality, Present)<br>(Certainty, Certain)<br>(Actor, Patient) | Event Classification:<br>The category of medication event "cholesterol-lowering agent" is "**Disposition**".<br><br>Context Classification:<br>The category of disposition event "cholesterol-lowering agent" from the dimension of Action is "**Start**". The category of disposition event "cholesterol-lowering agent" from the dimension of Negation is "**Negated**". The category of disposition event "cholesterol-lowering agent" from the dimension of Temporality is "**Present**". The category of disposition event "cholesterol-lowering agent" from the dimension of Certainty is "**Certain**". The category of disposition event "cholesterol-lowering agent" from the dimension of Actor is "**Patient**". |
| Progress note understanding | Assessment: "45 year old male with no known CAD and aspirin allergy who presented with chest pain and symptoms concerning for unstable angina, now with STE changes on ECG in the setting of chest pain. Now chest pain free."<br><br>Plan: "PUMP: Patient with some ECG changes mildly concerning for possible LVH, although does not meet diagnostic criteria on current ECG ' s. - Baseline TTE today to assess." | Direct | The relation between the given assessment and plan subsection is **Direct**. |

The ground-truth annotations are highlighted in bold.

**Evaluation metrics**

For clinical concept extraction and relation extraction, we used the strict micro-averaged precision, recall, and F1-score, aggregated across all concept and relation categories, for evaluation. For clinical concept normalization, we used both strict and relax precision, recall, and F1-score for evaluation. For clinical abbreviation disambiguation, we used strict precision, recall, and F1-score for evaluation. For NLI, we used accuracy for evaluation. For medication attribute filling and progress note understanding, we used F1-score for evaluation. We examined two GatorTronGPT models with 5 billion and 20 billion parameters and compared with previous transformer-based models.

**RESULTS**

The proposed soft prompting with a frozen generative LLM, GatorTronGPT, solved 7 clinical NLP tasks using a single unified model and achieved state-of-the-art performance on 5 out of 7 tasks. **Figure 2** summarizes the comparison results and **Table 2** provides detailed evaluation scores.

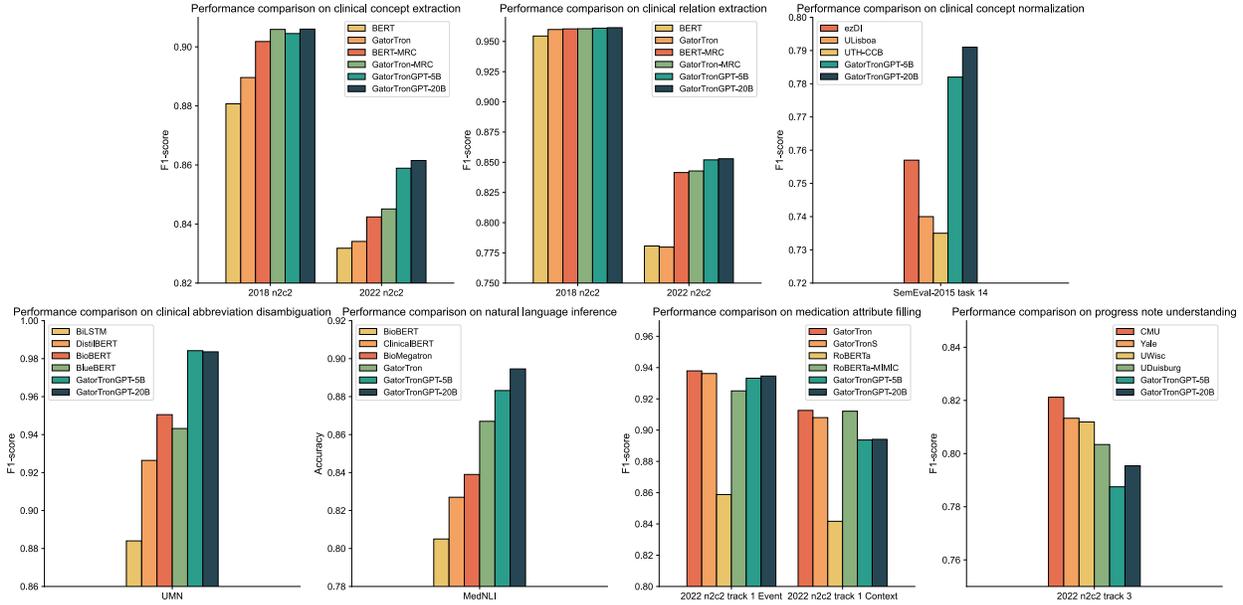

**Figure 2**. Summarization of comparison results on 7 clinical NLP tasks.

**Table 2**. Comparison of soft prompting using frozen GatorTronGPT with previous transformer-based models.

| 1. Clinical Concept Extraction | | | | | | |
|---|---|---|---|---|---|---|
| **Datasets** | **2018 n2c2** | | | **2022 n2c2** | | |
| **Model** | **Precision** | **Recall** | **F1** | **Precision** | **Recall** | **F1** |
| BERT | 0.8887 | 0.8728 | 0.8807 | 0.8160 | 0.8483 | 0.8318 |
| GatorTron | 0.8759 | 0.9038 | 0.8896 | 0.8181 | 0.8508 | 0.8341 |
| BERT-MRC | 0.9159 | 0.8942 | 0.9018 | 0.8496 | 0.8353 | 0.8424 |
| GatorTron-MRC | 0.9199 | 0.9012 | 0.9059 | 0.8521 | 0.8396 | 0.8451 |
| GatorTronGPT-5B | 0.9168 | 0.9057 | 0.9045 | 0.8598 | 0.8415 | 0.8589 |
| GatorTronGPT-20B | 0.9189 | 0.9081 | **0.9060** | 0.8615 | 0.8424 | **0.8615** |
| **2. Clinical Relation Extraction** | | | | | | |
| **Datasets** | **2018n2c2** | | | **2022 n2c2** | | |
| **Model** | **Precision** | **Recall** | **F1** | **Precision** | **Recall** | **F1** |
| BERT | 0.9598 | 0.9438 | 0.9545 | 0.7998 | 0.7620 | 0.7807 |
| GatorTron | 0.9719 | 0.9482 | 0.9599 | 0.7970 | 0.7631 | 0.7798 |
| BERT-MRC | 0.9722 | 0.9489 | 0.9604 | 0.8432 | 0.8371 | 0.8415 |
| GatorTron-MRC | 0.9724 | 0.9488 | 0.9605 | 0.8445 | 0.8390 | 0.8428 |
| GatorTronGPT-5B | 0.9734 | 0.9492 | 0.9610 | 0.8551 | 0.8472 | 0.8520 |
| GatorTronGPT-20B | 0.9746 | 0.9506 | **0.9614** | 0.8560 | 0.8491 | **0.8529** |
| **3. End-to-end Clinical Concept Extraction and Normalization** | | | | | | |
| **Dataset** | **SemEval-2015 task 14** | | | | | |
| **Model** | **Strict Precision** | **Strict Recall** | **Strict F1** | **Relax Precision** | **Relax Recall** | **Relax F1** |
| ezDI (best in challenge) | 0.783 | 0.732 | 0.757 | 0.815 | 0.761 | 0.787 |
| ULisboa | 0.779 | 0.705 | 0.740 | 0.806 | 0.729 | 0.765 |
| UTH-CCB | 0.778 | 0.696 | 0.735 | 0.797 | 0.714 | 0.753 |
| GatorTronGPT-5B | 0.810 | 0.751 | 0.782 | 0.833 | 0.769 | 0.795 |
| GatorTronGPT-20B | 0.812 | 0.772 | **0.791** | 0.839 | 0.785 | **0.813** |

| 4. Clinical Abbreviation Disambiguation | | | | 5. Natural Language Inference | |
|---|---|---|---|---|---|
| Dataset | UMN abbreviation | | | Dataset | MedNLI |
| Model | Precision | Recall | F1 | Model | Accuracy |
| BiLSTM | 0.8810 | 0.8960 | 0.8840 | BioBERT | 0.8050 |
| DistilBERT | 0.9254 | 0.9358 | 0.9263 | ClinicalBERT | 0.8270 |
| BioBERT | 0.9479 | 0.9597 | 0.9505 | BioMegatron | 0.8390 |
| BlueBERT | 0.9430 | 0.9518 | 0.9432 | GatorTron | 0.8670 |
| GatorTronGPT-5B | 0.9854 | 0.9832 | **0.9842** | GatorTronGPT-5B | 0.8832 |
| GatorTronGPT-20B | 0.9849 | 0.9830 | 0.9836 | GatorTronGPT-20B | **0.8946** |
| 6. Medication Attribute Filling. Dataset: 2022 n2c2 Track 1 | | | | 7. Progress Note Understanding | |
| | Event | Context | | Dataset | 2022 n2c2 |
| Model | F1 | F1 | | Model | F1 |
| GatorTron | **0.9379** | **0.9126** | | CMU (challenge best) | **0.8212** |
| GatorTronS | 0.9362 | 0.9080 | | Yale | 0.8133 |
| RoBERTa | 0.8588 | 0.8417 | | UWisc | 0.8119 |
| RoBERTa-MIMIC | 0.9251 | 0.9121 | | UDuisburg | 0.8034 |
| GatorTronGPT-5B | 0.9332 | 0.8937 | | GatorTronGPT-5B | 0.7875 |
| GatorTronGPT-20B | 0.9346 | 0.8941 | | GatorTronGPT-20B | 0.7954 |

CMU: Carnegie Mellon University; UWisc: University of Wisconsin; UDuisburg: University of Duisburg–Essen. Best evaluation scores are highlighted in bold.

**Clinical concept extraction.** The GatorTronGPT-20B model achieved state-of-the-art performance for both drug-ADE and SDoH datasets. For drug-ADE concept extraction, GatorTronGPT-20B outperformed task-specific models like BERT and GatorTron by 1.6-2.5%, achieved comparable performance to GatorTron-MRC, a machine reading comprehension architecture implemented using hard prompts based on an encoder-only LLM, GatorTron. For the SDoH dataset, which has more types of concepts and overlapping concepts, the GatorTronGPT-20B model outperformed task-specific models by ~3% and outperformed the GatorTron-MRC model by 1.6%. We observed performance improvements in both datasets when scaling up from the GatorTronGPT-5B to the GatorTronGPT-20B model.

**Clinical relation extraction.** The GatorTronGPT-20B model achieved state-of-the-art performance on both benchmark datasets. For drug-ADE relation, GatorTronGPT-20B outperformed task-specific BERT and GatorTron by 0.1 and 0.7%, respectively, and achieved comparable performance to GatorTron-MRC. For SDoH relation extraction, GatorTronGPT-20B outperformed other task-specific models by by ~7% and outperformed GatorTron-MRC by 1%.

**Clinical concept normalization**. **Table 2** compared GatorTronGPT-20B with the top 3 systems developed in this challenge. GatorTronGPT-20B achieved state-of-the-art performance outperforming the best system developed in the original challenge by 3.4% and 2.6% in strict and relax F1 scores, respectively.

**Clinical abbreviation disambiguation**. GatorTronGPT-5B achieved a state-of-the-art F1 score of 0.9842, outperforming previous task-specific transformer models by 3.4-10% on the UMN abbreviation disambiguation benchmark dataset.

**Natural language inference**. The GatorTronGPT-20B model achieved state-of-the-art accuracy, outperforming previous task-specific transformers by 5.5-9% and outperformed our previous GatorTron model by 2.8%.

**Medical attribute filling**. For the event classification task, GatorTronGPT-20B outperformed task-specific RoBERTa models by 0.9-7.6% and achieved comparable performance with our previously developed GatorTron and GatorTronS [5] – an encoder-only clinical LLM developed using synthetic clinical text generated by GatorTronGPT. For context classification, our previously developed GatorTron achieved the best performance.

**Progress note understanding**. GatorTronGPT-20B achieved performance comparable with the NLP model ranked 4[th] in this challenge, which is 2.6% lower than the best performance. The top 4 teams, including Carnegie Mellon University (CMU), Yale, University of Wisconsin (UWisc), and University of Duisburg–Essen (UDuisburg), are all based on an ensemble of multiple models [10]; whereas, the proposed method is a single model solution.

## DISCUSSION

Text-to-text learning through generative LLMs offers a unified solution across all clinical NLP tasks. Even though a previous study in the general English domain examined the encoder-decoder T5 model, their approach is still one step away from the promise of "one model for all NLP tasks". This study advances that work by exploring soft prompting with a frozen generative LLM, GatorTronGPT— developed from scratch using 277 billion words of mixed clinical and general English text with a GPT-3 architecture and up to 20 billion parameters—thereby fulfilling the goal of maintaining a single model from training to deployment. Our experiments show that the proposed approach achieved state-of-the-art performance on 5 out of 7 major clinical NLP tasks. The proposed solution also outperformed our previously developed encoder-only model GatorTron [30], and the hard prompt-based GatorTron-MRC models [29], demonstrating the versatility and efficiency of soft prompting with frozen generative LLMs in diverse clinical NLP tasks. We observed consistent performance improvements by scaling up the size of GatorTronGPT, indicating that large size generative LLMs enhance text-to-text learning. This study, along with our previous study[5], demonstrates that generative clinical LLMs can be all-purpose text analytics engines.

Our approach can directly benefit the real-world deployment of AI systems in healthcare. Real-world clinical decision support and healthcare require many complex AI modules, ranging from information extraction and standardization to various classifications and language understanding. In this study, we successfully solved 7 major clinical NLP tasks at the phrase-, sentence-, and document-levels, covering diverse tasks including information retrieval, classification, and language understanding. As more AI-enabled tools are being developed for healthcare, the burden of deploying and maintaining them grows significantly. The success of using a single generative LLM to solve these major clinical NLP tasks suggests that the proposed soft prompting with frozen generative LLMs holds promise for unifying diverse and complex AI modules in healthcare applications.

This study demonstrates that generative LLMs, such as GatorTronGPT, are more versatile than encoder-only LLMs such as BERT. Generative LLMs, designed for autoregressive text generation, use a decoder module to efficiently capture linguistic sequences and generate human-like responses. In generative LLMs, both input and output are sequences, which conveniently allows for formulating multiple NLP tasks using text-to-text learning. In contract, encoder-only transformer models are designed to derive vector representations of input tokens and require additional task-specific layers for different tasks.

To realize the "one model for all" concept from training to deployment, we adopted soft prompting with frozen generative LLMs. This study, along with our previous study [29] using encoder-only LLMs, shows that machines can learn "soft" prompts that are more effective and robust than "hard" prompts composed by human for both generative LLMs and encoder-only LLMs. Previous studies have reported that generative LLMs such as ChatGPT are very sensitive to hard prompts; even minor changes in prompts can lead to dramatically different responses. Our approach provides a robust approach enabling machines to learn robust soft prompts, freeing researchers from time-consuming task of prompt engineering. Frozen LLMs, i.e., keep LLM parameters unchanged during prompting, is critical to achieve "one model for all", as any fine-tuning of LLMs will generate task-specific variations deviating from the goal. We suspect that fine-tuning LLMs to a specific task will reduce their generalizability to other tasks. This can potentially be alleviated by tuning generative LLMs using many different tasks, i.e., multi-task instruction learning. Recent studies[43–45] have reported that multi-task instruction learning could further improve performance of generative LLMs.

Hallucination remains a challenge due to the probabilistic nature of text-to-text learning algorithms. We conducted an error analysis to examine the hallucinations made by GatorTronGPT. **Table 3** shows three types of hallucinations we observed including (1) Nonlogical, (2) Irrelevant, and (3) Interpretable. To ensure security and safety of healthcare, hallucinations must be controlled into a minimal level. Previously, we proposed a solution [29] to down-grade free text generation into a machine reading comprehension architecture, restricting LLMs to use only words presented in the given context for generating results. However, this solution is only applicable for information extraction and is not viable for other NLP tasks such as WSD and NLI. Our current approach, based on soft prompting with frozen generative LLMs, is generalizable to almost all major clinical NLP tasks.

**Table 3**. Observed hallucinations on the clinical relation extraction task.

|   | Input data | Ground truth | Hallucination |
|---|---|---|---|
| 1. Nonlogical hallucination: the LLM generated sequences without logic. | Disp: * 60 80 mg syringe * Refills: * 1 * 8. [s2] Acetaminophen [e2] 500 mg Tablet Sig: 1 - 2 Tablets PO [s1] Q6H (every 6 hours) as needed [e1] for pain. | The relation between "Q6H (every 6 hours) as needed" and "Acetaminophen" is **"Frequency-Drug"**. | - Non-responder  - Non-responder   - Non-responder |
|   | 8. docusate sodium 100 mg [s1] Capsule [e1] Sig: One (1) Capsule PO BID (2 times a day). 7. [s2] senna [e2] 8.6 mg Tablet Sig: One (1) Tablet PO BID (2 times a | The relation between "Capsule" and "senna" is **"No-relation"**. | 1: 2: 3: 4: 5: 6: 7: 8: 9: 10: 11: 12: 13: 14: |
| | day) as needed for constipation. | | |
| 2. Irrelevant hallucination: the LLM generated readable but irrelevant answers. | Disp: * 30 Tablet (s) * Refills: * 0 * 24. [s2] clopidogrel [e2] 75 mg Tablet Sig: [s1] One (1) [e1] Tablet PO DAILY (Daily) | The relation between "One (1)" and "clopidogrel" is "**Dosage-Drug**". | 1. The drug clopidogrel is a non-steroidal anti-platelet drug that is used to prevent platelet aggregation and reduce the risk of thrombosis in patients with acute coronary syndromes. It is also used to prevent thrombosis in patients with non-cardiac surgery. 2. Clopidogrel is a prodrug that is converted to its active form by the enzyme CYP2C19. |
| | Medications on Admission: atenolol 25 mg daily aspirin 81 mg daily lipitor 10 mg QOD (every other am) prednisone 10 mg daily tamsulosin SR 0.4 mg evening multivitamin 1 tab daily fish oil capsule 1000 mg twice a day [s2] systane lubricant eye [e2] drops 1 gtt [s1] TID [e1]. | The relation between "TID" and "systane lubricant eye" is "**Frequency-Drug**". | I have been using nonpreserved systane eye drops for a few years. |
| 3. Interpretable hallucination: the LLM generated a different way of the answer that is interpretable. | OXYCODONE - 20 mg Tablet Sustained Release 12 hr - 3 (Three) Tablet (s) by mouth every morning (60 mg), 1 tablet every 2 pm (20 mg) and 3 tablets every evening (60 mg) [s2] PANTOPRAZOLE [e2] [PROTONIX] - 40 mg Tablet , Delayed Release (E.C.) - [s1] 1 [e1] Tablet (s) by mouth once day | The relation between "1" and "PANTOPRAZOLE" is "**Dosage-Drug**". | The relation type between the Drug entity "PANTOPRAZOLE" and Dosage entity "1" is "has_dosage". |
| | Disp: * 30 Tablet (s) * Refills: * 2 * 3. [s2] Fluticasone - Salmeterol [e2] 250 - 50 mcg / Dose Disk with Device Sig: One (1) [s1] Disk with Device [e1] Inhalation Hospital 1 (2 times a day). | The relation between "Disk with Device" and "Fluticasone - Salmeterol" is "**Form-Drug**" | The relation type between the Drug entity "Fluticasone - Salmeterol" and Form entity "Disk with Device" is "is_a". |

# CONCLUSION

This study demonstrates that soft prompting with frozen generative LLMs is an all-purpose clinical text analytics engine solving various major clinical NLP tasks. Frozen LLM, i.e., keeping LLM parameters unchanged during prompt tuning, is the key to achieving "one unified model for all" solution. The proposed solution is cost-effective to adopt generative LLMs for various healthcare applications.

# ACKNOWLEDGMENTS

We would like to thank the i2b2 and n2c2 challenge organizers to provide the annotated corpus. We gratefully acknowledge the support of NVIDIA Corporation and the NIVIDA AI Technology Center (NVAITC) UF program. We would like to thank the UF Research Computing team, led by Dr. Erik Deumens, for providing computing power through UF HiperGator-AI cluster.

# FUNDING STATEMENT

This study was partially supported by a Patient-Centered Outcomes Research Institute® (PCORI®) Award (ME-2018C3-14754), a grant from the National Cancer Institute, R01CA246418, grants from the National Institute on Aging, NIA R56AG069880, R01AG080624, R01AG080624, R01AG083039, R01AG080991, R01AG084236,

R01AG084178, R01AG076234, and R33AG062884, Ed and Ethel Moore Alzheimer's Disease Research Program 23A09, National Institute of Allergy and Infectious Diseases, NIAID R01AI172875, the Cancer Informatics Shared Resource supported by the UF Health Cancer Center and the UF Clinical and Translational Science Institute Biomedical Informatics Program. The content is solely the responsibility of the authors and does not necessarily represent the official views of the funding institutions.

## COMPETING INTERESTS STATEMENT

KES, ABC, and MGF are employed by NVIDIA. There are no other competing financial or non-financial interests. The work presented in this study was conducted exclusively within the University of Florida Health.

## CONTRIBUTORSHIP STATEMENT

CP and YW were responsible for the overall design, development, and evaluation of this study. CP, AC, ZY, and KES had full access to all the data in the study and takes responsibility for the integrity of the data and the accuracy of the data analysis. CP and YW did the bulk of the writing, KES, ABC, MF, and JB also contributed to writing and editing of this manuscript. All authors reviewed the manuscript critically for scientific content, and all authors gave final approval of the manuscript for publication.

# DATA AVAILABILITY

The 2018 n2c2 and 2022 n2c2 datasets are available from the n2c2 website after data use agreement (see https://n2c2.dbmi.hms.harvard.edu/ for details).

The SemEval-2015 dataset is available from: https://alt.qcri.org/semeval2015/task14/.

The UMN abbreviation dataset is available from: https://conservancy.umn.edu/handle/11299/137703.

The MedNLI dataset is available from https://physionet.org/content/mednli/1.0.0/.